\documentclass[12pt]{article}
\usepackage[utf8]{inputenc}
\usepackage{geometry}
\usepackage[sortcites]{biblatex}
\bibliography{custom}
\usepackage{graphicx}
\usepackage{amsmath}
\usepackage{float}
\usepackage[table]{xcolor}
\usepackage{amssymb}
\usepackage{booktabs}
\usepackage{hyperref}

\title{Constructing Temporal Dynamic Knowledge Graphs from Interactive Text-based Games}
\author{Keunwoo Peter Yu}
\date{} 

\begin{document}
\maketitle

\begin{abstract}
In natural language processing, interactive text-based games serve as a test bed for interactive AI systems. Prior work has proposed to play text-based games by acting based on discrete knowledge graphs constructed by the Discrete Graph Updater (DGU) to represent the game state from the natural language description. While DGU has shown promising results with high interpretability, it suffers from lower knowledge graph accuracy due to its lack of temporality and limited generalizability to complex environments with objects with the same label. In order to address DGU's weaknesses while preserving its high interpretability, we propose the Temporal Discrete Graph Updater (TDGU), a novel neural network model that represents dynamic knowledge graphs as a sequence of timestamped graph events and models them using a temporal point based graph neural network. Through experiments on the dataset collected from a text-based game TextWorld, we show that TDGU outperforms the baseline DGU. We further show the importance of temporal information for TDGU's performance through an ablation study and demonstrate that TDGU has the ability to generalize to more complex environments with objects with the same label. All the relevant code can be found at \url{https://github.com/yukw777/temporal-discrete-graph-updater}.
\end{abstract}

\section{Introduction}

Text-based games are electronic games that employ text-based user interfaces. In contrast with video games that rely on visual feedback, a text-based game uses natural language to describe the state of the game and accept input from the user to interact with the game environment. More formally, a text-based game can be thought of as a partially observable environment in which an agent perceives and interacts with the environment purely through textual natural language \cite{ammanabrolu2021modeling}. In natural language processing, text-based games serve as a test bed for interactive AI systems that attempt to comprehend natural language and perform actions to interact with the environment as well as humans \cite{hausknecht2020interactive}.

Prior work that has attempted to effectively play text-based games has focused on constructing an accurate representation of the game state from the natural language descriptions in the form of a knowledge graph and then acting in the game based on the graph. Some attempts have focused on heuristics to build knowledge graphs, which are then used as the input to a deep learning model that selects the next action \cite{ammanabrolu2018playing,ammanabrolu2020graph}, while more recent work has proposed more general, data-driven approaches to building knowledge graphs \cite{adhikari-gata,zelinka2019building,ammanabrolu2021learning}. Particularly, \textcite{adhikari-gata} have proposed two models that incrementally update knowledge graphs based on textual observations that describe the game state instead of building them from scratch. The first model is the Continuous Graph Updater (CGU) that represents knowledge graphs as continuous and dense adjacency matrices updated as the hidden states of a recurrent neural network, making them uninterpretable. The second model is the Discrete Graph Updater (DGU) first proposed by \textcite{zelinka2019building} that updates knowledge graphs using ``graph update commands,'' in the form of object-relation-object RDF triples such as \texttt{add(table, kitchen, in)}. Thanks to this design choice, DGU produces interpretable knowledge graphs unlike CGU; however, DGU has resulted in worse game performance than CGU due to its lower knowledge graph accuracy caused by its lack of temporality, errors accumulating over game steps and errors caused by its discrete nature (e.g., round-off error). Furthermore, DGU's generalizability is limited as it cannot represent multiple objects with the same label due to label conflicts.

The goal of this work is to acquire interpretable knowledge graphs from interactive text-based games without sacrificing accuracy. We focus on constructing knowledge graphs, rather than choosing actions based on them, and on improving DGU so that it suffers less from errors while preserving its interpretability and improving its generalizability. To that end, we propose to replace the graph update commands with timestamped graph events and model them using a temporal point based graph neural network so that the model produces human-interpretable knowledge graphs that are more accurate and generalizable to environments with multiple objects with the same label.

We compare the performance of our model to DGU using the dataset provided by \textcite{adhikari-gata} and show that our model outperforms it. We also perform an ablation study to show the importance of modeling temporality in constructing dynamic knowledge graphs. Furthermore, we demonstrate that our model supports multiple objects with the same label by modifying the dataset to contain such objects and retraining our model on it. All the relevant code can be found at \url{https://github.com/yukw777/temporal-discrete-graph-updater}.

\section{Background and Related Work}
\subsection*{Interactive Text-based Games}
One of the major goals of AI research is to develop an agent that can seamlessly converse with humans and carry out appropriate actions. A key challenge to this goal is the fact that it is not scalable to test such agents directly with humans. Interactive text-based games have garnered interest in the AI research community as they can alleviate this scalability problem. There are two recent text-based game engines that have been designed for AI research: TextWorld \cite{cote2018textworld} and Jericho \cite{hausknecht2020interactive}, which researchers can use to generate interactive text-based games that involve spatial navigation. There are pre-generated collections of games available for each of the game engines: the First TextWorld Problems (FTWP) dataset\footnote{https://aka.ms/ftwp} and JerichoWorld \cite{ammanabrolu2021modeling}. There is another collection of games called LIGHT \cite{urbanek2019learning} that has been collected via crowdsourcing unlike procedurally generated FTWP and JerichoWorld. In this paper, we focus on TextWorld and the FTWP dataset as our baseline has used them.

\subsection*{Knowledge Representation and Knowledge Graphs}
As text-based games contain many distinct locations with different objects, finding a good knowledge representation is important to help the agent remember and focus on the most relevant information. This echos the Mental Model Theory \cite{johnson2010mental} from cognitive science, which states that humans reason by first constructing mental models. In robotics, this knowledge representation problem has been formalized as the simultaneous localization and mapping (SLAM) problem where a robot needs to incrementally build a consistent map of an unknown environment while simultaneously determining its location within the map \cite{durrant2006simultaneous}. Text-based games involve a variant of the SLAM problem, Textual-SLAM, where the agent needs to localize itself and build a model of its environment as it navigates and receives textual descriptions of an unknown environment \cite{ammanabrolu2021modeling}. \textcite{ammanabrolu2020graph,adhikari-gata,ammanabrolu2018playing,murugesan2020enhancing} have demonstrated that agents equipped with knowledge graphs as their knowledge representation perform better in text-based games than an end-to-end baseline that attempts to map textual observations directly to actions. In particular, \textcite{adhikari-gata} have established an upper bound on the performance of their agent by providing it with the ground truth knowledge graphs, showing the importance of the accuracy of constructed knowledge graphs. As a result, some works \cite{zelinka2019building,ammanabrolu2021learning} have focused on improving the accuracy of constructed knowledge graphs. In this work, we follow this line of research and propose a novel model that can construct interpretable knowledge graphs more accurately.

\subsection*{Dynamic Graph Neural Networks}
Knowledge graphs that are used to play text-based games are dynamic by nature as they need to be updated based on new textual observations and agent actions. However, traditional graph neural networks only handle static graphs. A simple way to circumvent this limitation is to represent a dynamic graph as a series of static graphs and use a traditional graph neural network in conjunction with a time series model like a recurrent neural network \cite{skardinga2021foundations}. On the other hand, STGCN \cite{yan2018spatial} transforms the series of static graphs into one static graph by connecting the related nodes across the series and uses a traditional graph neural network to model it instead of using a time series model. Another way to model a dynamic graph is to model the changes in graph structure as changes in edge weights. CN\textsuperscript{3} \cite{liu2019contextualized} calculates pair-wise node attention at each step to update edge weights, while Hybrid GNN \cite{liu2020retrieval} uses the structure-aware global attention mechanism.

The approaches described above do not apply to newly added or deleted nodes, which is crucial for text-based games as they represent new or removed objects. Furthermore, they use dense graph representations which are not scalable for a longer time horizon. In order to address these issues, researchers have recently proposed temporal point process based models that represent graphs as graph events from the graph stream representation, which can explicitly represent node and edge additions and deletions while staying sparse and compact even for a longer time horizon \cite{skardinga2021foundations}. In this work, we reference two of the latest temporal point process based models TGAT \cite{xu2020inductive} and TGN \cite{rossi2020temporal}.

\section{Dynamic Knowledge Graph Construction from Text}
In this section, we give details about the task of constructing dynamic knowledge graphs from text in text-based games. Specifically, we describe the way in which a dynamic knowledge graph can be built by repeatedly applying updates to it.

\subsection{Task Definition}
We follow the definition from \cite{zelinka2019building} where the full game state $s_t$ at any game step $t$ in a text-based game is represented by a knowledge graph $\mathcal{G}^{full}_t = (\mathcal{V}_t, \mathcal{E}_t)$. In TextWorld, nodes $\mathcal{V}_t$ represents entities (e.g., objects, the player and locations) and their states (e.g., closed, fried, sliced, etc.), while edges $\mathcal{E}_t$ that connect the nodes represent a set of relations between entities (e.g., north of, in, is, etc.).

\begin{figure}
    \centering
    \includegraphics[width=\textwidth]{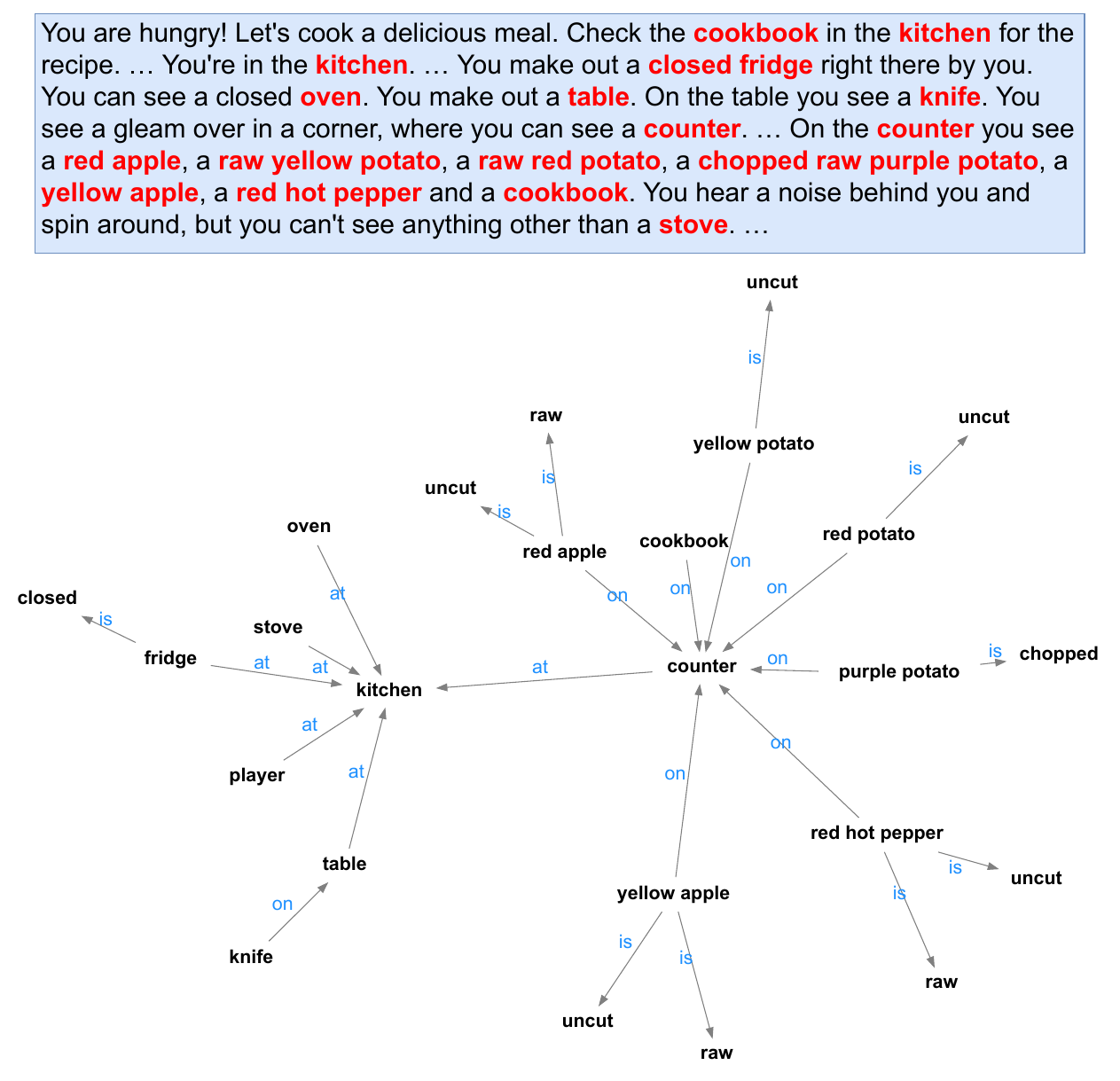}
    \caption{The ground truth seen belief graph given the textual observation. Emphasis in the textual observation is ours.}
    \label{fig:ground-truth-graph}
\end{figure}

Text-based games like TextWorld are partially observable; therefore, the agent does not have access to $\mathcal{G}^{full}_t$. Instead, it needs to build its own belief graph about the game state, $\mathcal{G}^{belief}_t$, based on textual observations. The goal of the agent should be to construct $\mathcal{G}^{belief}_t$ so that it matches the ground truth $\mathcal{G}^{seen}_t$, which is a subgraph of $\mathcal{G}^{full}_t$ that represents only the entities and relations that have been seen so far in the game. Figure \ref{fig:ground-truth-graph} shows an example of a ground truth $\mathcal{G}^{seen}_t$ based on a textual observation.

\subsection{Dynamic Knowledge Graph Construction} \label{dynamic-kg-construction}
\textcite{zelinka2019building} have proposed to construct dynamic knowledge graphs by representing updates to the agent's belief graph, $\Delta g_t$, using graph update commands such that $\mathcal{G}^{belief}_t = \texttt{U}(\mathcal{G}^{belief}_{t-1}, \Delta g_t)$ where \texttt{U} is an oracle function that applies $\Delta g_t$. They have defined two types of update commands:
\begin{itemize}
    \item \texttt{add(n1, n2, r)}: add a directed edge, labeled \texttt{r}, from \texttt{n1} to \texttt{n2}, while adding nodes that do not already exist.
    \item \texttt{delete(n1, n2, r)}: delete a directed edge, labeled \texttt{r}, from \texttt{n1} to \texttt{n2}. Ignore the command if any of the nodes or the edge does not exist.
\end{itemize}

The oracle function \texttt{U} does not apply $\Delta g_t$ until all the graph update commands have been generated. This means that each graph update command is generated without access to the latest knowledge graph, making the model prone to ``forgetting'' to add or delete nodes and edges. Also, the updated knowledge graph from \texttt{U} is a flat, static graph without any historical information about how the graph has been updated to the current state. This lack of temporality makes it difficult for the model to differentiate between similar nodes that have been added at different times, and prevents the model from exploiting useful biases such as recency bias (a new node is more likely to be attached to a recently added node than an old one). All of these limitations hamper the ability of the model to produce accurate knowledge graphs.

\textcite{zelinka2019building} have formulated the task of generating graph update commands into a classic token-based Seq2Seq problem where, given a textual observation $O_t$ and a previous belief graph $\mathcal{G}^{belief}_{t-1}$, the model generates a sequence of tokens that represent multiple update commands separated by a delimiter token. However, as these graph update commands are purely based on labels, they cannot represent more complex graphs that have multiple objects with the same label. For example, consider a case where the textual observation is ``There is an apple on the table. There is another apple on the chair.'' The resulting $\Delta g_t$, would be [\texttt{add(apple, table, on)}, \texttt{add(apple, chair, on)}], which results in an inaccurate graph where one apple exists both on the table and the chair simultaneously. Ideally, we want our graph to have two separate nodes for the two apples, each connected to the table and chair vertices separately.

We propose to address these limitations by replacing graph updated commands with timestamped graph events to represent the updates to the graph. First, we define timestamp \textbf{t} to be a two-dimensional vector $[t_g, t_e]$ where $t_g$ denotes the game step and $t_e$ denotes the graph event step within the game step. For example, the fifth graph event in the second game step would have \textbf{t} of $[1, 4]$ (zero-indexed). This two-dimensional timestamp vector allows us to handle the enhanced granularity provided by timestamped graph events. Following the definitions proposed by \textcite{rossi2020temporal} with the two-dimensional timestamp vector defined above, we define four types of timestamped graph events:

\begin{itemize}
    \item \texttt{node-add} event represented by $\textbf{v}_i(\textbf{t)}$ where $i$ denotes the index of the added node and $\textbf{v}$ is the attribute vector associated with the event.
    \item \texttt{node-delete} event represented by a tuple $(i, \textbf{t})$ where $i$ denotes the index of the deleted node.
    \item \texttt{edge-add} event represented by $\textbf{e}_{ij}(\textbf{t})$ where $i$ denotes the index of the source node, $j$ denotes the index of the destination node and $\textbf{e}$ is the attribute vector associated with the event.
    \item \texttt{edge-delete} event represented by a tuple $(i, j, \textbf{t})$ where $i$ denotes the index of the source node and $j$ denotes the index of the destination node of the deleted edge.
\end{itemize}

It is easy to see that timestamped graph events provide us with the flexibility to represent graphs with multiple objects with the same label. Given the textual observation from the earlier example, the resulting sequence of timestamped graph events would be [$\textbf{v}^{apple}_0([t_g, 0])$, $\textbf{v}^{table}_1([t_g, 1])$, $\textbf{e}^{on}_{01}([t_g, 2])$, $\textbf{v}^{apple}_2([t_g, 3])$, $\textbf{v}^{chair}_3([t_g, 4])$, $\textbf{e}^{on}_{23}([t_g, 5])$], where $\textbf{v}^l$ and $\textbf{e}^l$ represent the attribute vectors for a node and an edge with label $l$ respectively. Note that node 0 and 2 are both labeled ``apple,'' and yet they are represented by separate nodes, which is not possible when using graph update commands from \cite{zelinka2019building}.

Generating timestamped graph events can also be formulated as a Seq2Seq problem where given a textual observation $O_{t_g}$ and the current belief graph $\mathcal{G}^{belief}_\textbf{t}$, the model generates a sequence of timestamped graph events as structured prediction.

\begin{figure}
    \centering
    \includegraphics[width=\textwidth]{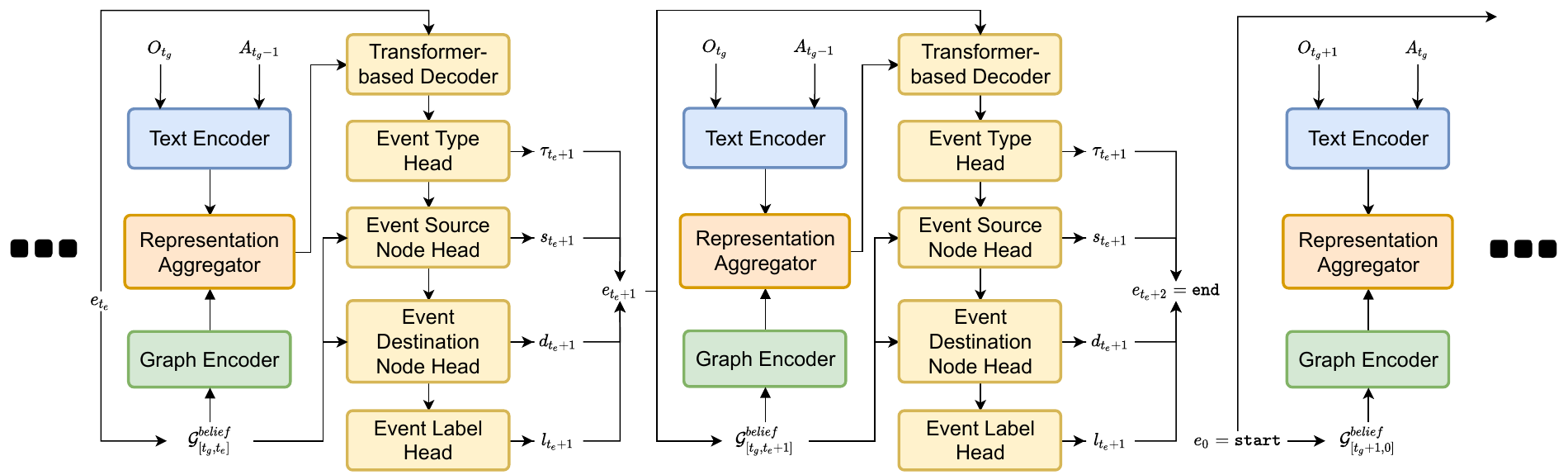}
    \caption{TDGU model architecture diagram that depicts the unrolled graph event generation process. The primary input to the model consists of the textual observation $O_{t_g}$ and the previous action $A_{t_g-1}$ that are encoded by the text encoder, and the current belief graph $\mathcal{G}^{belief}_{[t_g, t_e]}$ that is encoded by the graph encoder. The encoded representations are then aggregated by the representation aggregator, whose output is passed to the Transformer-based decoder alongside the graph event embedding. Finally, the decoder hidden state is given to the autoregressive heads that generate the next graph event $e_{t_e+1}=(\tau_{t_e+1}, s_{t_e+1}, d_{t_e+1}, l_{t_e+1})$, which is used to update $\mathcal{G}^{belief}_{[t_g, t_e]}$. The process continues until an \texttt{end} event is generated, at which point a \texttt{start} event, the textual input for the next game step, $O_{t_g+1}$ and $A_{t_g}$, and the current belief graph for the next game step $\mathcal{G}^{belief}_{[t_g+1, 0]}$ are given to the model to generate the next sequence of graph events.}
    \label{fig:model}
\end{figure}

\section{Temporal Discrete Graph Updater (TDGU)}
In this section, we introduce the Temporal Discrete Graph Updater (TDGU), a novel neural network model that updates discrete belief graphs based on textual observations. As shown in Figure \ref{fig:model}, the architecture consists of four main modules:
\begin{enumerate}
    \item A \textbf{text encoder} which encodes the textual observation $O_{t_g}$ and the action taken in the previous game step $A_{t_g-1}$.
    \item A \textbf{graph encoder} which encodes the current belief graph $\mathcal{G}^{belief}_\textbf{t}$.
    \item A \textbf{representation aggregator} which aggregates the encoded representations from the text encoder and the graph encoder.
    \item A \textbf{graph event decoder} which takes the aggregated representation and generates graph events.
\end{enumerate}

\subsection{Text Encoder}
TDGU uses the same Transformer-based \cite{vaswani2017attention} encoder used by \textcite{zelinka2019building,adhikari-gata}, which consists of a word embedding layer and a Transformer block, as its text encoder. TDGU's text encoder encodes the current textual observation $O_{t_g}$ and the previous action $A_{t_g-1}$ separately to produce context-dependent representations $\textbf{H}^{O_{t_g}} \in \mathbb{R}^{L^{O_{t_g}} \times H}$ and $\textbf{H}^{A_{t_g-1}} \in \mathbb{R}^{L^{A_{t_g-1}} \times H}$ respectively, where $L^{O_{t_g}}$ is the length of $O_{t_g}$, $L^{A_{t_g-1}}$ is the length of $A_{t_g-1}$ and $H$ is the dimension of the encoded representations:

\begin{align}
    \textbf{H}^{O_{t_g}} &= \texttt{TransEnc}([\texttt{Lin}_{text}(\texttt{WordEmb}(O^0_{t_g}))+\texttt{Pos}_{text}(0); ...; \nonumber \\
    & \qquad \texttt{Lin}_{text}(\texttt{WordEmb}(O^{L^{O_{t_g}}}_{t_g}))+\texttt{Pos}_{text}(L^{O_{t_g}})]) \\
    \textbf{H}^{A_{t_g-1}} &= \texttt{TransEnc}([\texttt{Lin}_{text}(\texttt{WordEmb}(A^0_{t_g-1}))+\texttt{Pos}_{text}(0); ...; \nonumber \\
    & \qquad \texttt{Lin}_{text}(\texttt{WordEmb}(A^{L^{A_{t_g-1}}}_{t_g-1}))+\texttt{Pos}_{text}(A_{t_g-1})])
\end{align}
$O^i_{t_g}$ and $A^i_{t_g-1}$ refer to the $i$'th token for $O_{t_g}$ and $A_{t_g-1}$ respectively. $\texttt{WordEmb}$ is the word embedding layer, which is initialized using the 300-dimensional fastText \cite{bojanowski2016enriching} word embeddings pretrained on Common Crawl (600B tokens). $\texttt{Lin}_{text}$ is a linear layer that reduces the dimension of the word embeddings to $H$, which is the dimension of the input and output of the Transformer block. $[\cdot;\cdot]$ is the vector concatenation operator. \texttt{TransEnc} is the Transformer encoder block that consists of a stack of five convolutional layers and a self-attention layer, followed by two linear layers with a ReLU non-linear activation layer in between. $\texttt{Pos}_{text}(i) \in \mathbb{R}^H$ is the positional encoding \cite{vaswani2017attention} for position $i$.

\subsection{Graph Encoder}
We first formally define the \texttt{node-add} event $\textbf{v}^l_i(\textbf{t)} \in \mathbb{R}^{H+H_{temp}}$ and the \texttt{edge-add} event $\textbf{e}^l_{ij}(\textbf{t)} \in \mathbb{R}^{H+H_{temp}}$:

\begin{align}
    \textbf{v}^l_i(\textbf{t}) &= [\frac{1}{L^{l}}\sum_{w \in l} \texttt{Lin}_{text}(\texttt{WordEmb}(w)); \texttt{Pos}_{graph}(t_g); \texttt{Pos}_{graph}(t_e)] \label{eq:node-attr} \\
    \textbf{e}^l_{ij}(\textbf{t}) &= [\frac{1}{L^{l}}\sum_{w \in l} \texttt{Lin}_{text}(\texttt{WordEmb}(w)); \texttt{Pos}_{graph}(t_g); \texttt{Pos}_{graph}(t_e)] \label{eq:edge-attr}
\end{align}
where \textbf{t} denotes the current timestamp $[t_g, t_e]$, $l$ is the label of length $L^l$ and $\texttt{Pos}_{graph}(i) \in \mathbb{R}^{H_{temp}/2}$ is the positional encoding for position $i$. Note that we use positional encoding as the temporal embeddings, which is a departure from some of the latest temporal point based graph neural networks such as TGAT and TGN. Their temporal embedding scheme relies on harmonic analysis in order to capture the relative time difference. However, absolute timestamps are more important in TextWorld given its incremental nature and shorter time horizon, which is better captured by positional encoding.

We then dynamically construct a node attribute matrix $V_\textbf{t} \in \mathbb{R}^{N_\textbf{t}^V \times (H+H_{temp})}$ and an edge attribute matrix $E_\textbf{t} \in \mathbb{R}^{N_\textbf{t}^E \times (H+H_{temp})}$ where $N_\textbf{t}^V$ and $N_\textbf{t}^E$ are the number of nodes and the number of edges at timestamp $\textbf{t}$. Specifically, for each \texttt{node-add} event $\textbf{v}^l_i(\textbf{t})$, we add a row to $V_\textbf{t}$ at position $i$ with the attribute vector as defined in Equation \ref{eq:node-attr}, and for each \texttt{node-delete} event $(i, \textbf{t})$, we remove row $i$ from $V_\textbf{t}$. Edge events are handled similarly with respect to $E_\textbf{t}$.

Once $V_\textbf{t}$ and $E_\textbf{t}$ are updated, we use an off-the-shelf graph neural network to obtain the final node embeddings $\textbf{H}^\mathcal{G}_\textbf{t} \in \mathbb{R}^{N_\textbf{t}^V \times H}$. Specifically,

\begin{equation}
    \textbf{H}^\mathcal{G}_\textbf{t} = \texttt{Lin}_{graph}(\texttt{ReLU}(\texttt{TransConv}(V_\textbf{t}, E_\textbf{t})))
\end{equation}
where \texttt{TransConv} is a one-headed graph Transformer operator \cite{shi2020masked}, \texttt{ReLU} is the ReLU activation layer and $\texttt{Lin}_{graph}$ is a linear layer that reduces the dimension of its input from $H+H_{temp}$ to $H$.

\subsection{Representation Aggregator}
TDGU uses the same representation aggregator used by \textcite{zelinka2019building,adhikari-gata} to aggregate text and graph representations. Specifically, we obtain an aggregated observation-to-graph representation $\textbf{H}^{O\mathcal{G}}_\textbf{t} \in \mathbb{R}^{L^{O_{t_g}} \times H}$ by first calculating a similarity matrix $\textbf{S} \in \mathbb{R}^{L^{O_{t_g}} \times N_\textbf{t}^V}$:

\begin{equation}
    \textbf{S} = \texttt{Sim}(\textbf{H}^{O_{t_g}}, \textbf{H}^\mathcal{G}_\textbf{t})
\end{equation}
where \texttt{Sim} is a trilinear similarity function \cite{seo2016bidirectional}. By applying softmax along both axes of \textbf{S}, we obatin $\textbf{S}^\mathcal{G}$ and $\textbf{S}^{O_{t_g}}$. Finally, $\textbf{H}^{O\mathcal{G}}_\textbf{t}$ is calculated:

\begin{align}
    \textbf{P} &= \textbf{S}^\mathcal{G}\textbf{H}^{\mathcal{G}}_\textbf{t} \\
    \textbf{Q} &= \textbf{S}^{O_{t_g}}\textbf{S}^{O_{t_g}\intercal}\textbf{H}^{O_{t_g}} \\
    \textbf{H}^{O\mathcal{G}}_\textbf{t} &= \texttt{Lin}_{aggr}([\textbf{H}^{O_{t_g}}; \textbf{P}; \textbf{H}^{O_{t_g}} \odot \textbf{P}; \textbf{H}^{O_{t_g}} \odot \textbf{Q}])
\end{align}
where $\texttt{Lin}_{aggr}$ is a linear layer that reduces the input dimension from $4H$ to $H$ and $\odot$ is the element-wise multiplication operator.

An aggregated graph-to-observation representation $\textbf{H}^{\mathcal{G}O}_\textbf{t} \in \mathbb{R}^{N_\textbf{t}^V \times H}$ can be calculated similarly. $\textbf{H}^{A\mathcal{G}}_\textbf{t} \in \mathbb{R}^{L^{A_{t_g-1}} \times H}$ and $\textbf{H}^{\mathcal{G}A}_\textbf{t} \in \mathbb{R}^{N_\textbf{t}^V \times H}$ are calculated the same way between $\textbf{H}^{A_{t_g-1}}$ and $\textbf{H}^\mathcal{G}_\textbf{t}$.

\subsection{Graph Event Decoder}
We formulate the task of generating a sequence of timestamped graph events as a Seq2Seq problem and use an autoregressive Transformer-based decoder similar to the one used by \textcite{zelinka2019building,adhikari-gata}. The input to the decoder is a sequence of timestamped graph event embeddings, which are constructed by concatenating the learned event type embedding, the mean projected word embeddings of the source node, destination and event labels. Furthermore, the aggregated representations from the representation aggregator are given to the decoder to attend over. The output of the decoder then goes through a series of autoregressive heads, similar to the ones proposed by \textcite{vinyals2019grandmaster}, that are designed to handle structured and combinatorial timestamped graph events. There are four autoregressive heads: the event type head, the event source node head, the event destination node head and the event label head. A new graph event is generated based on the predictions made by the four heads, which is then appended to the sequence of timestamped graph events and fed back to the decoder to generate the next graph event.

\subsubsection{Transformer-based Decoder}
In order to formulate timestamped graph event generation as a Seq2Seq problem, we first represent $e_i$, the $i$th graph event in the sequence generated so far, as a tuple of four arguments $(\tau_i, s_i, d_i, l_i)$ where $\tau_i$ denotes the event type, $s_i$ denotes the source node, $d_i$ denotes the destination node, and $l_i$ denotes the event label. Timestamp information is not included in this tuple, as $t_g$ is encoded in the aggregated representations and $t_e$ is given to the decoder via positional encoding. Each argument in the tuple is then transformed into an embedding and concatenated to produce a graph event embedding $\textbf{g}_i \in \mathbb{R}^{H_\tau+ 3H}$:

\begin{align}
    \textbf{g}_i &= [\texttt{EventTypeEmb}(\tau_i); \nonumber \\
                 & \quad \frac{1}{L^{l_{s_i}}}\sum_{w \in l_{s_i}} \texttt{Lin}_{text}(\texttt{WordEmb}(w)); \nonumber \\
                 & \quad \frac{1}{L^{l_{d_i}}}\sum_{w \in l_{d_i}} \texttt{Lin}_{text}(\texttt{WordEmb}(w)); \nonumber \\
                 & \quad \frac{1}{L^{l_i}}\sum_{w \in l_i} \texttt{Lin}_{text}(\texttt{WordEmb}(w))]
\end{align}
where \texttt{EventTypeEmb} is the learned event type embedding layer with dimension $H_\tau$, $l_{s_i}$ is the label of the source node and $l_{d_i}$ is the label of the destination node. Following the standard practice for Seq2Seq models, we add two additional special event types, \texttt{start} and \texttt{end}, to the event type vocabulary. Furthermore, as certain event types do not require all the arguments, we appropriately mask out parts of $\textbf{g}_i$: for the special event types, we mask all the embeddings except for the event type embedding; for \texttt{node-add}, we mask the embeddings for the source and destination nodes; for \texttt{node-delete}, we mask the embeddings for the destination node and the event label; for \texttt{edge-add}, we do not apply any masking; for \texttt{edge-delete}, we mask the embedding for the event label.

We pass the sequence of graph event embeddings as well as the aggregated representations to a Transformer-based decoder with masked attention to generate a hidden vector $\textbf{h}^{dec}_\textbf{t} \in \mathbb{R}^H$:

\begin{align}
    \textbf{h}^{dec}_\textbf{t} &= \texttt{TransDec}([\texttt{Lin}_{dec}(\textbf{g}_0)+\texttt{Pos}_{dec}(0); ...; \texttt{Lin}_{dec}(\textbf{g}_{t_e})+\texttt{Pos}_{dec}(t_e)], \nonumber \\
    & \qquad \qquad \qquad \textbf{H}^{O\mathcal{G}}_\textbf{t}, \textbf{H}^{\mathcal{G}O}_\textbf{t}, \textbf{H}^{A\mathcal{G}}_\textbf{t}, \textbf{H}^{\mathcal{G}A}_\textbf{t})
\end{align}
where $\texttt{Lin}_{dec}$ is a linear layer that reduces the dimension of the input from $H_\tau + 3H$ to $H$ and \texttt{TransDec} is the Transformer-based decoder block, which consists of a self-attention layer, a multihead-attention layer for each of the aggregated representation, followed by two linear layers with a ReLU activation layer in between.

\subsubsection{Event Type Head}
The event type head is the first of the autoregressive heads as the event type determines which graph event argument is necessary. It calculates a distribution over the event types $P(\mathcal{T} = \tau|O_{t_g},A_{t_g-1},\mathcal{G}^{belief}_\textbf{t})$ that can be used to generate the next event type. Specifically,

\begin{align}
    \textbf{h}^{\tau(0)}_{\textbf{t}} &= \texttt{ReLU}(\texttt{LN}(\texttt{Lin}^0_\tau(\textbf{h}^{dec}_\textbf{t})))) \\
    \textbf{h}^{\tau(1)}_\textbf{t} &= \texttt{ReLU}(\texttt{LN}(\texttt{Lin}^1_\tau(\textbf{h}^{\tau(0)}_\textbf{t}))) \\
    P(\mathcal{T}|O_{t_g},A_{t_g-1},\mathcal{G}^{belief}_\textbf{t}) &= \texttt{Softmax}(\texttt{Lin}^2_\tau(\textbf{h}^{\tau(1)}_\textbf{t})) \\
    \tau_{t_e+1} &= \operatorname{argmax}_i P(\mathcal{T}=i|O_{t_g},A_{t_g-1},\mathcal{G}^{belief}_\textbf{t})
\end{align}
where \texttt{LN} is layer normalization and $\texttt{Lin}_\tau^j$ is the $j$th linear layer for the event type head.

The event type head also creates an autoregressive embedding $\textbf{h}_\textbf{t}^{auto(\tau)} \in \mathbb{R}^{H_{auto}}$ from $\textbf{h}^{dec}_\textbf{t}$ and $\tau_{t_e+1}$:

\begin{equation}
    \textbf{h}_\textbf{t}^{auto(\tau)} = \texttt{Lin}_\tau^{dec}(\textbf{h}^{dec}_\textbf{t}) + \texttt{Lin}_\tau^{auto(1)}(\texttt{ReLU}(\texttt{Lin}_\tau^{auto(0)}(\texttt{OneHot}(\tau_{t_e+1}))))
\end{equation}
where \texttt{OneHot} is a function that returns a one-hot vector, $\texttt{Lin}_\tau^{dec}$ is a linear layer that maps $\textbf{h}^{dec}_\textbf{t}$ to the autoregressive embedding space and $\texttt{Lin}_\tau^{auto(i)}$ is the $i$the linear layer to map the one-hot event type vector to the autoregressive embedding space.

\subsubsection{Event Source Node Head}
The event source node head uses the query-key attention mechanism to calculate a distribution over the nodes $P(S=s|O_{t_g},A_{t_g-1},\mathcal{G}^{belief}_\textbf{t},\tau_{t_e+1})$ where $S$ is the set of possible source nodes, which we assume to be all nodes $\mathcal{V}_\textbf{t}$, in $\mathcal{G}^{belief}_\textbf{t}$. Specifically, it calculates a key matrix $\textbf{K} \in \mathbb{R}^{N_\textbf{t}^V \times H_{node}}$ and a query vector $\textbf{q} \in \mathbb{R}^{H_{node}}$:

\begin{align}
    \textbf{K} &= \texttt{Conv1D}(\textbf{H}^\mathcal{G}_\textbf{t}) \\
    \textbf{q} &= \texttt{Lin}_{src}^1(\texttt{ReLU}(\texttt{Lin}_{src}^0(\textbf{h}_\textbf{t}^{auto(\tau)})))
\end{align}
where \texttt{Conv1D} is a 1D convolution layer and $\texttt{Lin}_{src}^i$ is the $i$th linear layer that maps $\textbf{h}_\textbf{t}^{auto(\tau)}$ to the query-key space. It then multiplies $\textbf{K}$ and $\textbf{q}$ together to calculate the distribution over the nodes:

\begin{gather}
    P(S|O_{t_g},A_{t_g-1},\mathcal{G}^{belief}_\textbf{t},\tau_{t_e+1}) = \texttt{Softmax}(\textbf{K}\textbf{q}) \\
    s_{t_e+1} = \operatorname{argmax}_i P(S=i|O_{t_g},A_{t_g-1},\mathcal{G}^{belief}_\textbf{t},\tau_{t_e+1})
\end{gather}

The event source node head then creates an autoregressive embedding $\textbf{h}_\textbf{t}^{auto(s)} \in \mathbb{R}^{H_{auto}}$ based on $\textbf{K}$ and $\textbf{h}_\textbf{t}^{auto(\tau)}$:

\begin{align}
    \bar{\textbf{k}} &= (\frac{1}{N_\textbf{t}^V}\sum_i\textbf{K}_i)^\intercal \\
    \textbf{h}_\textbf{t}^{auto(s)} &= \textbf{h}_\textbf{t}^{auto(\tau)} + \texttt{Lin}_{src}^{auto}(\textbf{K}^\intercal\texttt{OneHot}(s_{t_e+1})-\bar{\textbf{k}})
\end{align}
where $\textbf{K}_i$ denotes the $i$th row of $\textbf{K}$ and $\texttt{Lin}_{src}^{auto}$ is a linear layer that maps from the query-key space to the autoregressive embedding space.

\subsubsection{Event Destination Node Head}
The event destination node head uses the same architecture as the event source node head to calculate a distribution over the nodes $P(D=d|O_{t_g},A_{t_g-1},\mathcal{G}^{belief}_\textbf{t},\tau_{t_e+1},s_{t_e+1})$, except it takes $\textbf{h}_\textbf{t}^{auto(s)}$ instead of $\textbf{h}_\textbf{t}^{auto(\tau)}$ as its input. We also assume that the set of possible destination nodes $D$ is $\mathcal{V}_\textbf{t}$. The event destination node then calculates an autoregressive embedding $\textbf{h}_\textbf{t}^{auto(d)} \in \mathbb{R}^{H_{auto}}$ the same way using the selected destination node $d_{t_e+1}$.

\subsubsection{Event Label Head}
The event label head uses $\textbf{h}_\textbf{t}^{auto(d)}$ to calculate a distribution over the label vocabulary $P(L=l|O_{t_g},A_{t_g-1},\mathcal{G}^{belief}_\textbf{t},\tau_{t_e+1},s_{t_e+1},d_{t_e+1})$:

\begin{align}
    \textbf{h}^{l(0)}_{\textbf{t}} &= \texttt{ReLU}(\texttt{LN}(\texttt{Lin}^0_l(\textbf{h}_\textbf{t}^{auto(d)})))) \\
    \textbf{h}^{l(1)}_\textbf{t} &= \texttt{ReLU}(\texttt{LN}(\texttt{Lin}^1_l(\textbf{h}^{l(0)}_\textbf{t}))) \\
    P(L&|O_{t_g},A_{t_g-1},\mathcal{G}^{belief}_\textbf{t},\tau_{t_e+1},s_{t_e+1},d_{t_e+1}) = \texttt{Softmax}(\texttt{Lin}^2_l(\textbf{h}^{l(1)}_\textbf{t})) \\
    l_{t_e+1} &= \operatorname{argmax}_i P(L=i|O_{t_g},A_{t_g-1},\mathcal{G}^{belief}_\textbf{t},\tau_{t_e+1},s_{t_e+1},d_{t_e+1})
\end{align}
Since this is the last autoregressive head, the autoregressive embedding is not updated. Having gone through all the autoregressive heads, the next graph event $e_{t_e+1}$ can now be generated as $(\tau_{t_e+1}, s_{t_e+1}, d_{t_e+1}, l_{t_e+1})$.

\section{Experiments}
\subsection{Dataset}
We use the command generation dataset provided by \textcite{adhikari-gata}, which they have used to train DGU. This is an updated version of the dataset provided by \textcite{zelinka2019building}. The command generation dataset was collected from the games in the FTWP dataset by taking the difference between two consecutive ground truth seen knowledge graphs. Specifically, an agent follows the ``walkthrough'' steps for each game, which are the most efficient steps to beat the game provided by TextWorld, while additionally taking 10 random steps at each walkthrough step. At each game step, differences between the ground truth seen knowledge graphs are calculated and encoded as graph update commands described in Section \ref{dynamic-kg-construction}. As a result, each datapoint in the command generation dataset contains the textual observation, previous action, previous ground truth seen knowledge graph and the target update commands. Table \ref{tab:dataset-stat} shows the basic statistics for the command generation dataset.

\begin{table}
\centering
\begin{tabular}{@{}cccccccc@{}}
\toprule
Train & Valid & Test & Avg. Obs. & Avg. Cmds. & Nodes & Edges & Avg. Conns. \\ \midrule
413455  & 20177   & 64749  & 29.3      & 2.67              & 99         & 10      & 30.47              \\ \bottomrule
\end{tabular}
\caption{Statistics of the command generation dataset. \textbf{Avg. Obs.} is the average number of tokens for textual observations. \textbf{Avg. Cmds.} is the average number of graph update commands. \textbf{Nodes} and \textbf{Edges} refer to the number of node and edge types. \textbf{Avg. Conns.} is the average number of edges per graph.}
\label{tab:dataset-stat}
\end{table}

\subsection{Preprocessing}
We preprocess the command generation dataset by turning the graph update commands into timestamped graph events. We first sort the update commands the same way \textcite{adhikari-gata,zelinka2019building} have done. Then for each datapoint, the previous ground truth seen knowledge graph and the target graph update commands are each transformed into a sequence of timestamped graph events by progressively building a graph using the graph update commands. During this process, exit nodes and state nodes need to be handled properly as they need to be treated as separate nodes despite having the same label due to the limitations in expressiveness of the update commands described in Section \ref{dynamic-kg-construction}. For example, a room in TextWorld can have multiple exits, and multiple food items can be sliced. The textual observation and previous action are tokenized using spaCy\footnote{https://spacy.io/} as \textcite{adhikari-gata,zelinka2019building} have done. As a result, each preprocessed datapoint contains the tokenized textual observation and previous action, the current ground truth seen knowledge graph as a sequence of timestamped graph events and the target graph events.

\subsection{Training Setup}
TDGU is trained via teacher-forcing where the model is given the ground truth sequence of timestamped graph events as its input. Specifically, the graph encoder receives the current ground truth seen knowledge graph in the form of a sequence of timestamped graph events, and each of the autoregressive heads receives the previous ground truth event argument, i.e. the source node head receives the ground truth event type; the destination node head receives the ground truth source node; and the event label head receives the ground truth destination node. The text encoder simply receives the preprocessed textual observation and previous action.

We calculate a negative log-likelihood loss for each autoregressive head. The total loss is a weighted sum of all the losses from the heads. We use the weighting strategy proposed by \textcite{kendall2018multi} with a small modification proposed by \textcite{liebel2018auxiliary} to calculate the weights for the losses. Table \ref{tab:hyperparam} shows the various hyperparameters of TDGU. The model was trained for 2.5 days on a NVIDIA A40 GPU.

\begin{table}
\centering
\begin{tabular}{@{}ll@{}}
\toprule
Hyperparameter         & Value            \\ \midrule
$H$                    & 64               \\
$H_{temp}$             & 16               \\
$H_\tau$               & 16               \\
$H_{auto}$             & 128              \\
$H_{node}$             & 16               \\
Batch size             & 64               \\
Learning rate          & $5\times10^{-4}$ \\
Optimization algorithm & AdamW            \\ \bottomrule
\end{tabular}
\caption{Various hyperparameters for TDGU training.}
\label{tab:hyperparam}
\end{table}

\subsection{Experimental Setup and Results}
\subsubsection{Dynamic Knowledge Graph Construction}
We first test the effectiveness of TDGU by comparing its performance in dynamic knowledge graph construction to DGU. For fair comparison with the baseline, we follow the evaluation strategy proposed by \textcite{zelinka2019building} and measure the teacher-force (TF) F\textsubscript{1} score, as well as the free-run (FR) F\textsubscript{1} score. These scores are based on the intersection of the generated set of graph update commands or \textit{RDF} triples and the ground truth set.

\begin{itemize}
    \item \textbf{TF F\textsubscript{1}}: the model uses greedy decoding to generate timestamped graph events based on the current ground truth belief graph $\mathcal{G}^{belief}_\textbf{t}$.  As \textcite{zelinka2019building} measured this score based on graph update commands, we convert the model's generated timestamped graph events into graph update commands. Specifically, we take the event type, the source node label, destination node label and event label of a timestamped edge event to create a graph update command. Then we calculate the F\textsubscript{1} score using the set of converted graph update commands and the ground truth update commands.
    \item \textbf{FR F\textsubscript{1}}: Unlike \textcite{zelinka2019building} who measured this score per game, we calculate this score per ``trajectory'', which consists of walkthrough steps and random steps. For example, given a game with 5 walkthrough steps, the second trajectory would consist of the first two walkthrough steps and 10 random steps taken from the second walkthrough step. For each trajectory, we start with an empty belief graph and update it using greedy decoding without using any ground truth belief graph until the end of the trajectory. Then we convert each edge in the final belief graph into an \textit{RDF} triple and calculate the F\textsubscript{1} score against the final ground truth belief graph. This is the metric that is most representative of the actual performance of the model in realistic settings.
\end{itemize}

\begin{figure}
    \centering
    \includegraphics[width=\textwidth]{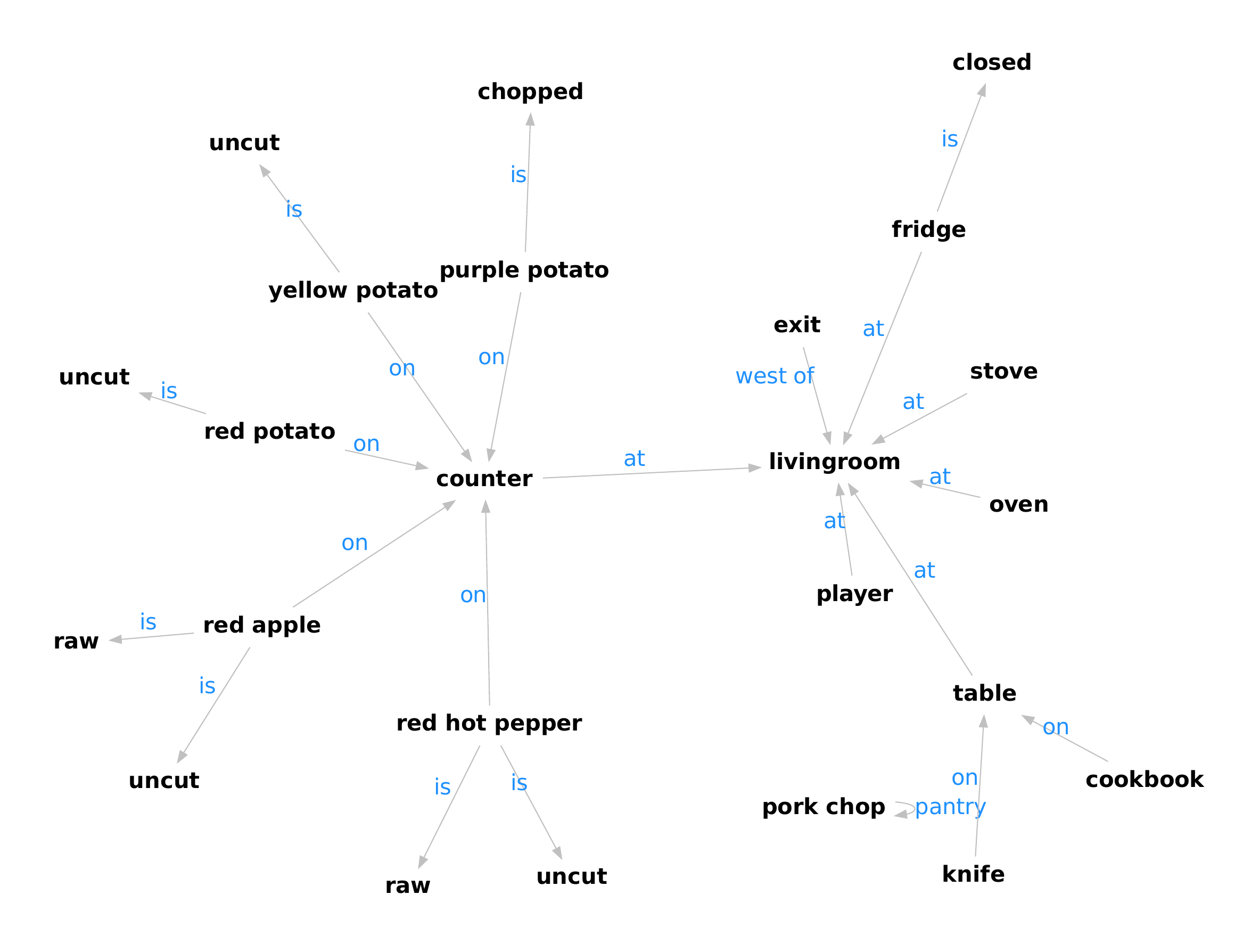}
    \caption{A graph constructed by TDGU based on the same textual observation from Figure \ref{fig:ground-truth-graph}. It is mostly accurate but has some glaring issues such as the mislabeled ``kitchen'' node (labeled as ``livingroom''), misattached ``cookbook'' node, missing ``yellow apple'', and the self-looped ``pork chop'' node with a ``pantry'' edge.}
    \label{fig:tdgu-graph}
\end{figure}

\begin{figure}
    \centering
    \includegraphics[width=\textwidth]{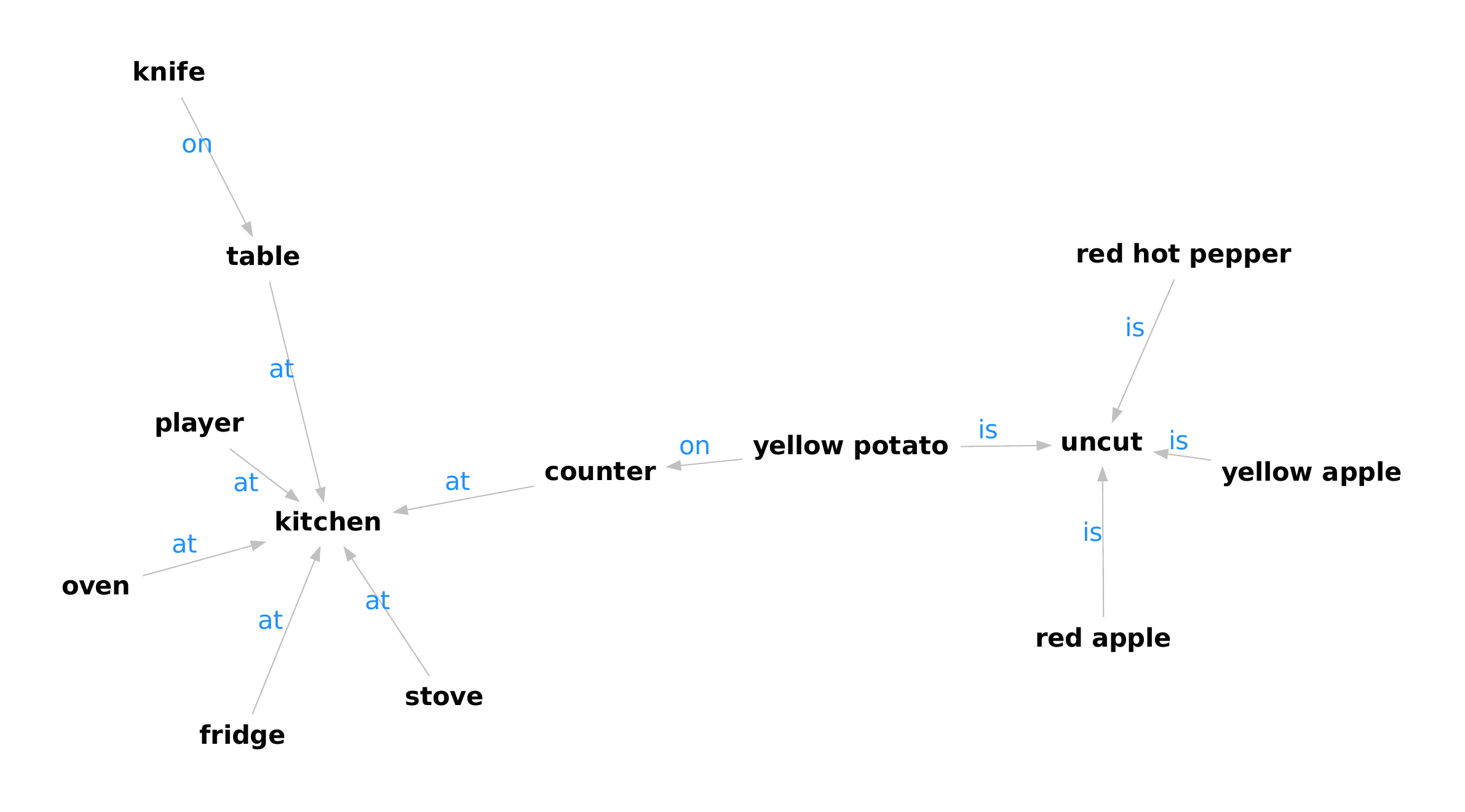}
    \caption{A graph constructed by DGU based on the same textual observation from Figure \ref{fig:ground-truth-graph}. Unlike TDGU (Figure \ref{fig:tdgu-graph}), DGU correctly labeled the ``kitchen'' node. However, it missed the ``closed'' node that should be attached to the ``fridge'' node. Furthermore, it struggled with placing food items correctly ``on'' the ``counter''. As described in Section \ref{dynamic-kg-construction}, it only created one ``uncut'' node to which all the food item nodes are attached, instead of creating one for each of the uncut food items.}
    \label{fig:dgu-graph}
\end{figure}

\textcite{zelinka2019building} have only reported the scores on the older version of the command generation dataset and have not released their code. The code released by \textcite{adhikari-gata} only reports TF F\textsubscript{1} on the updated version of the command generation dataset. In order to establish a fair comparison, we retrain DGU using their released code on the updated command generation dataset, and use the same evaluation code we wrote for TDGU to calculate FR F\textsubscript{1}. Results are reported in Table \ref{tab:results}. Figure \ref{fig:tdgu-graph} and \ref{fig:dgu-graph} show example graphs constructed by TDGU and the baseline DGU respectively.

\begin{table}
\centering
\begin{tabular}{@{}lll@{}}
\toprule
                                     & TF F\textsubscript{1} & FR F\textsubscript{1} \\ \midrule
DGU (baseline)                       & \textbf{0.969}        & 0.782                 \\
TDGU                                 & 0.916                 & \textbf{0.849}        \\
TDGU\textsubscript{\texttt{no-temp}} & 0.912                 & 0.797                 \\
TDGU\textsubscript{\texttt{multi}}   & N/A                   & 0.823                 \\ \bottomrule
\end{tabular}
\caption{TF F\textsubscript{1} and FR F\textsubscript{1} scores of TDGU and its variants, as well as the baseline DGU.}
\label{tab:results}
\end{table}

\subsubsection{Modeling Temporality}
\begin{figure}
    \centering
    \includegraphics[width=\textwidth]{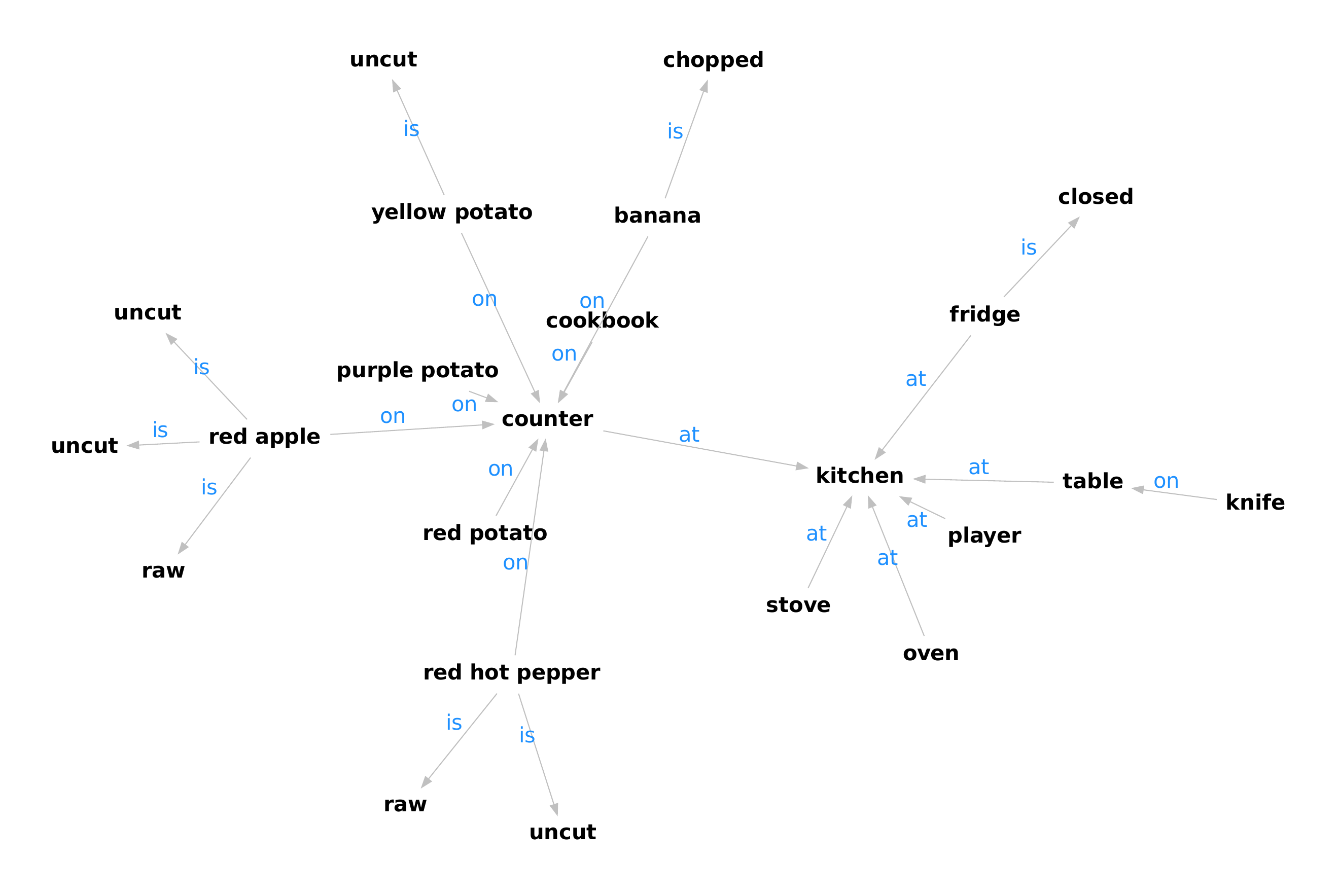}
    \caption{A graph constructed by TDGU\textsubscript{\texttt{no-temp}} based on the same textual observation from Figure \ref{fig:ground-truth-graph}. Unlike TDGU (Figure \ref{fig:tdgu-graph}), TDGU\textsubscript{\texttt{no-temp}} correctly labeled the ``kitchen'' node. However, it also missed the ``yellow apple'' node and added an erroneous ``banna'' node. Most critically, it struggled with attaching nodes with the same label (e.g., ``uncut'') to the correct nodes, which TDGU did not struggle with. This is because the embeddings of these nodes become very similar without the temporal embeddings, and TDGU\textsubscript{\texttt{no-temp}} struggles to differentiate between them.}
    \label{fig:tdgu-no-temp-graph}
\end{figure}

We train a version of TDGU, TDGU\textsubscript{\texttt{no-temp}}, with static (zero) positional encoding in order to gauge the importance of modeling temporality. The same preprocessed command generation dataset is used to train TDGU\textsubscript{\texttt{no-temp}}. The performance of TDGU\textsubscript{\texttt{no-temp}} is also reported in Table \ref{tab:results}. Figure \ref{fig:tdgu-no-temp-graph} shows an example graphs constructed by TDGU\textsubscript{\texttt{no-temp}}.

\subsubsection{Multiple Objects with Same Label}
In order to demonstrate TDGU's flexibility in handling multiple objects with the same label, we train another version of TDGU, TDGU\textsubscript{\texttt{multi}}, on a modified version of the command generation dataset that contains multiple objects with the same label. TextWorld does not natively support multiple objects with the same label within a game. However, it does contain food items in different colors such as yellow potatoes and purple potatoes. We split the nodes for these colored food items into two, one for the food item and the other for the color, and connect them with an edge labeled ``is''. For example, if the textual observation is ``There is a purple potato on the table. There is a yellow potato on the chair.'', the resulting graph events would be [$\textbf{v}^{potato}_0([t_g, 0])$, $\textbf{v}^{purple}_1([t_g, 1])$, $\textbf{e}^{is}_{01}([t_g, 2])$, $\textbf{v}^{table}_2([t_g, 3])$, $\textbf{e}^{on}_{02}([t_g, 4])$, $\textbf{v}^{potato}_{3}([t_g, 5])$, $\textbf{v}^{yellow}_4([t_g, 6])$, $\textbf{e}^{is}_{34}([t_g, 7])$, $\textbf{v}^{chair}_5([t_g, 8])$, $\textbf{e}^{on}_{35}([t_g, 9])$].

\begin{figure}
    \centering
    \includegraphics[width=\textwidth]{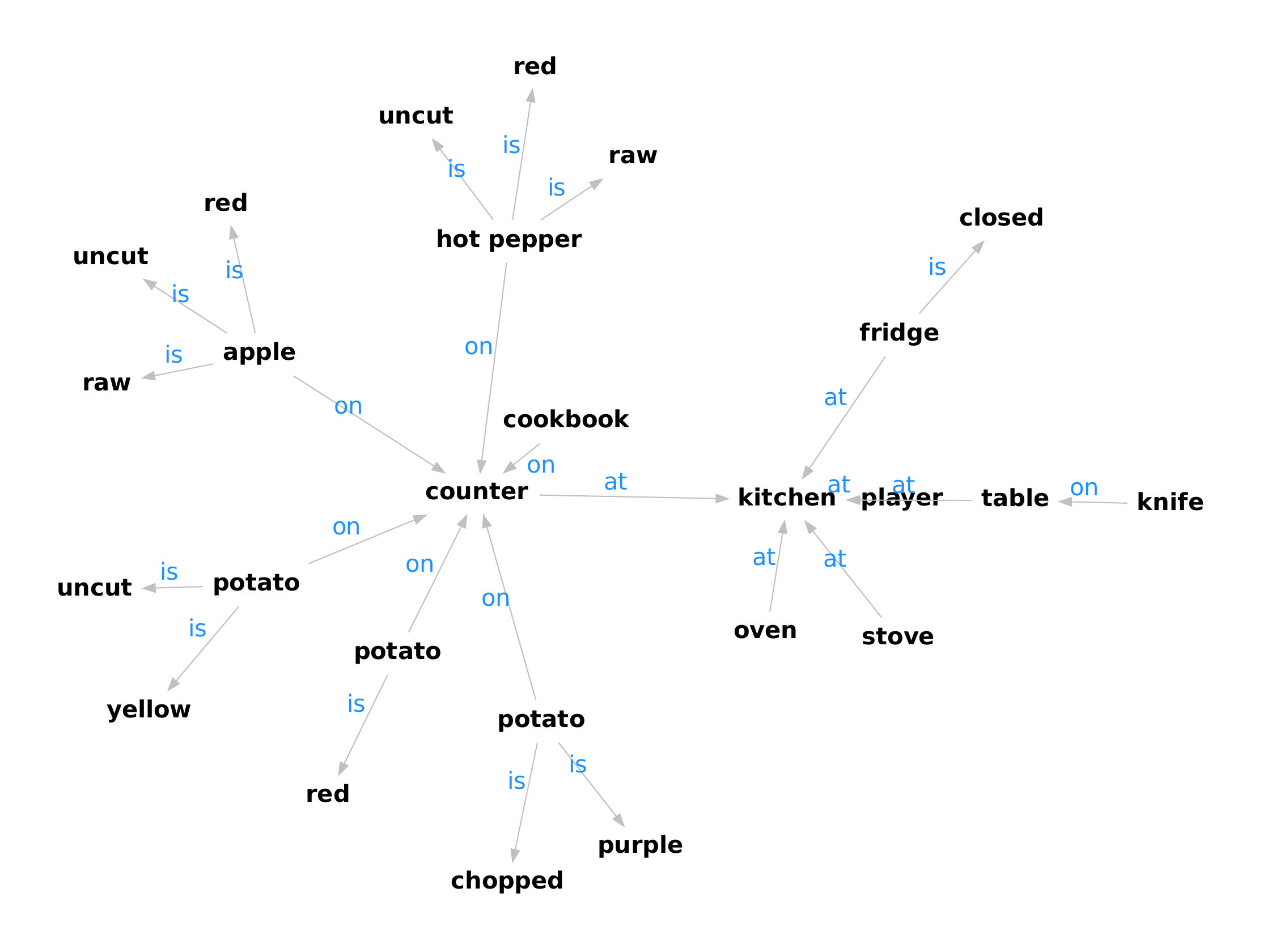}
    \caption{A graph constructed by TDGU\textsubscript{\texttt{multi}} based on the textual observation from Figure \ref{fig:ground-truth-graph}. It performed the best out of all the models on this example, only missing the ``yellow apple'' node. Note that it handled objects with the same label, e.g. ``potato'', correctly.}
    \label{fig:tdgu-objs-same-label}
\end{figure}

TF F\textsubscript{1} cannot be measured in this setting as the graph update commands cannot represent updates for multiple objects with the same label. \textit{RDF} triples also lack the ability to represent such objects, but we were able to circumvent this limitation and measure FR F\textsubscript{1} by merging the nodes of colored food items into one when generating \textit{RDF} triples, and comparing them with the original ground truth \textit{RDF} triples. The FR F\textsubscript{1} score for this experiment is also reported in Table \ref{tab:results}. Figure \ref{fig:tdgu-objs-same-label} shows an example of graphs with multiple objects with the same label generated by TDGU.

\section{Discussion}
TDGU outperformed the baseline DGU in terms of FR F\textsubscript{1}. This is evidence that TDGU has better performance in realistic settings where ground truth knowledge graphs are not provided. TDGU\textsubscript{\texttt{no-temp}} performed worse than TDGU but slightly outperformed the baseline in terms of FR F\textsubscript{1}, which shows that the extra temporal information is indeed helpful in accurately updating knowledge graphs.

Both TDGU and TDGU\textsubscript{\texttt{no-temp}} performed worse than the baseline in terms of TF F\textsubscript{1}. We hypothesize that it might be due to the fact that the extra temporal information is a distraction when generating graph events for one game step based on ground truth knowledge graphs. The fact that both TDGU and TDGU\textsubscript{\texttt{no-temp}} have similar TF F\textsubscript{1} scores may be another piece of evidence that for short horizons, i.e. one game step, the extra temporal information may not help. We leave the investigation into this phenomenon for future studies.

The FR F\textsubscript{1} score of TDGU on the modified dataset with multiple objects with the same label was not as high as its performance on the original dataset mostly due to the fact that the modified dataset is harder as TDGU has to learn to differentiate between nodes with the same label. Even so, it outperformed the baseline while properly handling multiple objects with the same label as shown in Figure \ref{fig:tdgu-objs-same-label}.

\section{Future Directions}
There are a few architectural improvements we can make to TDGU to enhance its performance and generalizability. The first is adding an event edge head to the graph event decoder. Currently, TDGU makes separate classifications for the source node and destination node of the deleted edges. These classifications are autoregressive, but an event edge head that directly classifies which edge should be deleted can provide a larger architectural inductive bias, thereby increasing TDGU's performance. Furthermore, we can replace the current event label head, which is a simple classification layer, with an autoregressive decoder with the same vocabulary as the text encoder. This would allow TDGU to be more general in terms of the labels it can use for nodes and edges and easily be used in more complex environments beyond TextWorld. Moreover, we could add ``extractiveness'' to the model by using the pointer-generator network \cite{see2017get} to further enhance the accuracy of the event label head. It would also be interesting to try other graph neural networks for the graph encoder to see if they would have a positive impact on the model performance.

Upon implementing the architectural improvements described above, we plan to test TDGU on other benchmarks such as LIGHT and Jericho. It would also be interesting to see if using pretrained language models could enhance TDGU's performance on these benchmarks, especially LIGHT which should have more variability in textual descriptions as it was crowdsourced.

As graph events can have any kind of attribute vectors, they can be used to represent dynamic graphs with more complex node and edge features. One interesting future direction in this regard would be to use TDGU in multi-modal settings where node and edge features are formed from text as well as vision. This multi-modal model can use a pretrained large language model and vision model as its encoders in order to handle natural language and images better. 

Another interesting future direction is to explore ways to train TDGU in a self-supervised way. Currently, one of the biggest challenges in building a knowledge graph construction model like TDGU is the dearth of training data. If we can introduce appropriate architectural changes to TDGU so that it can be trained without the ground truth knowledge graphs, we can use TDGU for more complex tasks like commonsense reasoning where the inductive bias provided by the graphical nature of TDGU would be very helpful.

\section{Conclusion}
In this work, we have presented a novel neural network model that constructs dynamic knowledge graphs using timestamped graph events, and showed that it outperforms the baseline on a text-based game knowledge graph dataset in terms of the FR F\textsubscript{1} score, which is more representative of the performance in realistic settings. Furthermore, we have shown that our model is more generalizable to more complex environments than the baseline, specifically the ones with multiple objects with the same label; and therefore, it may open up opportunities to use dynamic knowledge graphs in tasks other than text-based games.

\printbibliography

@inproceedings{adhikari-gata,
 author = {Adhikari, Ashutosh and Yuan, Xingdi and C\^{o}t\'{e}, Marc-Alexandre and Zelinka, Mikul\'{a}\v{s} and Rondeau, Marc-Antoine and Laroche, Romain and Poupart, Pascal and Tang, Jian and Trischler, Adam and Hamilton, Will},
 booktitle = {Advances in Neural Information Processing Systems},
 editor = {H. Larochelle and M. Ranzato and R. Hadsell and M. F. Balcan and H. Lin},
 pages = {3045--3057},
 publisher = {Curran Associates, Inc.},
 title = {Learning Dynamic Belief Graphs to Generalize on Text-Based Games},
 url = {https://proceedings.neurips.cc/paper/2020/file/1fc30b9d4319760b04fab735fbfed9a9-Paper.pdf},
 volume = {33},
 year = {2020}
}

@article{zelinka2019building,
  title={Building dynamic knowledge graphs from text-based games},
  author={Zelinka, Mikul{\'a}{\v{s}} and Yuan, Xingdi and C{\^o}t{\'e}, Marc-Alexandre and Laroche, Romain and Trischler, Adam},
  journal={arXiv preprint arXiv:1910.09532},
  year={2019}
}

@article{ammanabrolu2021learning,
  title={Learning Knowledge Graph-based World Models of Textual Environments},
  author={Ammanabrolu, Prithviraj and Riedl, Mark O},
  journal={arXiv preprint arXiv:2106.09608},
  year={2021}
}

@article{ammanabrolu2020graph,
  title={Graph constrained reinforcement learning for natural language action spaces},
  author={Ammanabrolu, Prithviraj and Hausknecht, Matthew},
  journal={arXiv preprint arXiv:2001.08837},
  year={2020}
}

@article{ammanabrolu2018playing,
  title={Playing text-adventure games with graph-based deep reinforcement learning},
  author={Ammanabrolu, Prithviraj and Riedl, Mark O},
  journal={arXiv preprint arXiv:1812.01628},
  year={2018}
}

@article{murugesan2020enhancing,
  title={Enhancing text-based reinforcement learning agents with commonsense knowledge},
  author={Murugesan, Keerthiram and Atzeni, Mattia and Shukla, Pushkar and Sachan, Mrinmaya and Kapanipathi, Pavan and Talamadupula, Kartik},
  journal={arXiv preprint arXiv:2005.00811},
  year={2020}
}

@article{durrant2006simultaneous,
  title={Simultaneous localization and mapping: part I},
  author={Durrant-Whyte, Hugh and Bailey, Tim},
  journal={IEEE robotics \& automation magazine},
  volume={13},
  number={2},
  pages={99--110},
  year={2006},
  publisher={IEEE}
}

@article{skardinga2021foundations,
  title={Foundations and modelling of dynamic networks using dynamic graph neural networks: A survey},
  author={Skardinga, Joakim and Gabrys, Bogdan and Musial, Katarzyna},
  journal={IEEE Access},
  year={2021},
  publisher={IEEE}
}

@article{rossi2020temporal,
  title={Temporal graph networks for deep learning on dynamic graphs},
  author={Rossi, Emanuele and Chamberlain, Ben and Frasca, Fabrizio and Eynard, Davide and Monti, Federico and Bronstein, Michael},
  journal={arXiv preprint arXiv:2006.10637},
  year={2020}
}

@inproceedings{vaswani2017attention,
  title={Attention is all you need},
  author={Vaswani, Ashish and Shazeer, Noam and Parmar, Niki and Uszkoreit, Jakob and Jones, Llion and Gomez, Aidan N and Kaiser, {\L}ukasz and Polosukhin, Illia},
  booktitle={Advances in neural information processing systems},
  pages={5998--6008},
  year={2017}
}

@article{bojanowski2016enriching,
  title={Enriching Word Vectors with Subword Information},
  author={Bojanowski, Piotr and Grave, Edouard and Joulin, Armand and Mikolov, Tomas},
  journal={arXiv preprint arXiv:1607.04606},
  year={2016}
}

@article{xu2020inductive,
  title={Inductive representation learning on temporal graphs},
  author={Xu, Da and Ruan, Chuanwei and Korpeoglu, Evren and Kumar, Sushant and Achan, Kannan},
  journal={arXiv preprint arXiv:2002.07962},
  year={2020}
}

@article{shi2020masked,
  title={Masked label prediction: Unified message passing model for semi-supervised classification},
  author={Shi, Yunsheng and Huang, Zhengjie and Feng, Shikun and Zhong, Hui and Wang, Wenjin and Sun, Yu},
  journal={arXiv preprint arXiv:2009.03509},
  year={2020}
}

@article{vinyals2019grandmaster,
  title={Grandmaster level in StarCraft II using multi-agent reinforcement learning},
  author={Vinyals, Oriol and Babuschkin, Igor and Czarnecki, Wojciech M and Mathieu, Micha{\"e}l and Dudzik, Andrew and Chung, Junyoung and Choi, David H and Powell, Richard and Ewalds, Timo and Georgiev, Petko and others},
  journal={Nature},
  volume={575},
  number={7782},
  pages={350--354},
  year={2019},
  publisher={Nature Publishing Group}
}

@inproceedings{kendall2018multi,
  title={Multi-task learning using uncertainty to weigh losses for scene geometry and semantics},
  author={Kendall, Alex and Gal, Yarin and Cipolla, Roberto},
  booktitle={Proceedings of the IEEE conference on computer vision and pattern recognition},
  pages={7482--7491},
  year={2018}
}

@article{liebel2018auxiliary,
  title={Auxiliary tasks in multi-task learning},
  author={Liebel, Lukas and K{\"o}rner, Marco},
  journal={arXiv preprint arXiv:1805.06334},
  year={2018}
}

@article{see2017get,
  title={Get to the point: Summarization with pointer-generator networks},
  author={See, Abigail and Liu, Peter J and Manning, Christopher D},
  journal={arXiv preprint arXiv:1704.04368},
  year={2017}
}

@inproceedings{cote2018textworld,
  title={Textworld: A learning environment for text-based games},
  author={C{\^o}t{\'e}, Marc-Alexandre and K{\'a}d{\'a}r, Akos and Yuan, Xingdi and Kybartas, Ben and Barnes, Tavian and Fine, Emery and Moore, James and Hausknecht, Matthew and El Asri, Layla and Adada, Mahmoud and others},
  booktitle={Workshop on Computer Games},
  pages={41--75},
  year={2018},
  organization={Springer}
}

@article{ammanabrolu2021modeling,
  title={Modeling Worlds in Text},
  author={Ammanabrolu, Prithviraj and Riedl, Mark O},
  journal={arXiv preprint arXiv:2106.09578},
  year={2021}
}

@inproceedings{hausknecht2020interactive,
  title={Interactive fiction games: A colossal adventure},
  author={Hausknecht, Matthew and Ammanabrolu, Prithviraj and C{\^o}t{\'e}, Marc-Alexandre and Yuan, Xingdi},
  booktitle={Proceedings of the AAAI Conference on Artificial Intelligence},
  volume={34},
  number={05},
  pages={7903--7910},
  year={2020}
}

@article{seo2016bidirectional,
  title={Bidirectional attention flow for machine comprehension},
  author={Seo, Minjoon and Kembhavi, Aniruddha and Farhadi, Ali and Hajishirzi, Hannaneh},
  journal={arXiv preprint arXiv:1611.01603},
  year={2016}
}

@article{johnson2010mental,
  title={Mental models and human reasoning},
  author={Johnson-Laird, Philip N},
  journal={Proceedings of the National Academy of Sciences},
  volume={107},
  number={43},
  pages={18243--18250},
  year={2010},
  publisher={National Acad Sciences}
}

@article{urbanek2019learning,
  title={Learning to speak and act in a fantasy text adventure game},
  author={Urbanek, Jack and Fan, Angela and Karamcheti, Siddharth and Jain, Saachi and Humeau, Samuel and Dinan, Emily and Rockt{\"a}schel, Tim and Kiela, Douwe and Szlam, Arthur and Weston, Jason},
  journal={arXiv preprint arXiv:1903.03094},
  year={2019}
}

@inproceedings{yan2018spatial,
  title={Spatial temporal graph convolutional networks for skeleton-based action recognition},
  author={Yan, Sijie and Xiong, Yuanjun and Lin, Dahua},
  booktitle={Thirty-second AAAI conference on artificial intelligence},
  year={2018}
}

@inproceedings{liu2019contextualized,
  title={Contextualized non-local neural networks for sequence learning},
  author={Liu, Pengfei and Chang, Shuaichen and Huang, Xuanjing and Tang, Jian and Cheung, Jackie Chi Kit},
  booktitle={Proceedings of the AAAI Conference on Artificial Intelligence},
  volume={33},
  number={01},
  pages={6762--6769},
  year={2019}
}

@inproceedings{liu2020retrieval,
  title={Retrieval-Augmented Generation for Code Summarization via Hybrid GNN},
  author={Liu, Shangqing and Chen, Yu and Xie, Xiaofei and Siow, Jing Kai and Liu, Yang},
  booktitle={International Conference on Learning Representations},
  year={2020}
}

\end{document}